\documentclass[runningheads]{llncs}
\usepackage{graphicx}
\usepackage{subcaption}
\usepackage{wrapfig}
\usepackage{xcolor}
\usepackage{array}
\usepackage{float}
\usepackage{theorem}
\usepackage{textfit}
\usepackage{placeins}
\usepackage{framed, color}
\usepackage{url}
\usepackage{rotating}
\usepackage{times}
\usepackage{amsmath}
\usepackage{amsfonts}
\usepackage{algpseudocode}
\usepackage{algorithm}
\usepackage{tabularx}
\usepackage{booktabs}
\usepackage{ragged2e}
\usepackage{listings}
\usepackage[breaklinks]{hyperref}
\usepackage{tabto}
\usepackage{lipsum}
\usepackage[export]{adjustbox}
\usepackage{caption}
\usepackage{multirow}
\usepackage{enumitem}
\usepackage{xcolor}
\usepackage[justification=centering]{caption}

\usepackage{listings}
\usepackage{orcidlink}

\begin{document}
\title{A Hybrid AI and Rule-Based Decision Support System for Disease Diagnosis and Management Using Labs}

\titlerunning{Hybrid Clinical Decision Support System}
%
\author{Muhammad Hammad Maqsood\inst{1}\orcidlink{0009-0000-2983-0487}  \and Mubashir Sajid\inst{1}\orcidlink{0009-0003-5050-9542} \and Khubaib Ahmed\inst{1}\orcidlink{0009-0001-0562-9874} \and Muhammad Usamah Shahid\inst{1}\orcidlink{0009-0001-4293-2979} \and
Muddassar Farooq\inst{1}}
\authorrunning{H. Maqsood, M. Sajid, K. Ahmed, M. U. Shahid, M. Farooq}
%
\institute{CureMD Research, 80 Pine St 21st Floor, New York, NY 10005, United States
\email{\{hammad.maqsood, mubashir.sajid, khubaib.ahmed, muhammad.usamah, muddassar.farooq\}@curemd.com}\newline
\url{https://www.curemd.com/}}
\maketitle              

\begin{abstract}
This research paper outlines the development and
implementation of a novel Clinical Decision Support System (CDSS) that integrates AI predictive modeling with medical knowledge bases. It utilizes the quantifiable information elements in lab results for inferring likely diagnoses a patient might have. Subsequently, suggesting investigations to confirm the likely diagnoses -- an assistive tool for physicians.
The system fuses knowledge contained in a rule-base expert system with inferences of data driven predictors based on the features in labs. The data for 593,055 patients was collected from 547 primary care centers across the US to model our decision support system and derive Real-Word Evidence (RWE) to make it relevant for a large demographic of patients.
Our Rule-Base comprises clinically validated rules, modeling 59 health conditions that can directly confirm one or more of diseases and assign ICD-10 codes to them. The Likely Diagnosis system uses multi-class classification, covering 37 ICD-10 codes, which are grouped together into 11 categories based on the labs that physicians prescribe to confirm the diagnosis. This research offers a novel system that assists a physician by utilizing medical profile of a patient and routine lab investigations to predict a group of likely diseases and then confirm them, coupled with providing explanations for inferences, thereby assisting physicians to reduce misdiagnosis of patients in clinical decision-making.
\end{abstract}

\section{Introduction}
Clinical Decision Support Systems (CDSS) are vital tools in modern healthcare, designed to enhance medical decisions by leveraging knowledge and patients' data to provide insights. These systems help physicians use RWE derived from Real World EMR Data.
These systems integrate data from a variety of sources to recommend actionable interventions, improving patient outcomes and streamlining healthcare processes. They assist healthcare providers by offering decision-support tools that are directly integrated into clinical workflows of EMR systems. CDSS have been shown to improve valued based outcome for patients by reducing the cost of healthcare services. CDSS tools have been proposed that either utilize predictive modeling or draw inference using the rules in a medical rule base system \cite{sutton2020overview}. 

Despite advances in technology, many clinical decision support systems (CDSS) remain restricted, primarily utilizing either rule-based algorithms or classical artificial intelligence (AI) techniques without fully integrating both approaches to harness their combined potential\cite{chen2023harnessing}. 

Knowledge driven rule-based system have historically enjoyed higher acceptability among physicians as the rules can be clinically validated by providers. They incorporate domain knowledge directly into the system and assist in the decision making of a physician. In these systems, modelling knowledge representation and upgrading continuously is a significant challenge; as a result, gaps in care holes pop in these systems \cite{kong2009applying} with advancements in medical knowledge and associated technologies.

Machine learning (ML) and big data analytics have enhanced the sophistication of CDSS systems. Modern CDSS systems leverage ML models to offer predictive analytics and patient centered recommendations \cite{berge2023machine} \cite{chen2023harnessing}. These ML algorithms are able to achieve higher accuracy for a number of chronic diseases. Clinicians are, however, reluctant to trust these systems because of their black box type nature of making inferences.

The trust and acceptability of Predictive ML models in healthcare is increasing \cite{amann2020explainability}, especially as assistive tools, for the models that also explain their inference method. In this paper, we make efforts to add explainability layer on all ML tools. The major contributions of this research include: (1) the fusion of Knowledge Base with ML Models to design a Hybrid CDSS; (2) Utilization of quantifiable information in Laboratory Data for designing Likely diagnoses models for patients; and (3) Explanation of inferences to a physician using SHAP values to take informed decisions with the assistance of AI models.

\section{The Proposed CDSS}
\subsection{System Overview}
\begin{figure}[ht!]
    \centering
    \begin{minipage}{0.48\textwidth}
        \centering
        \includegraphics[width=\linewidth]{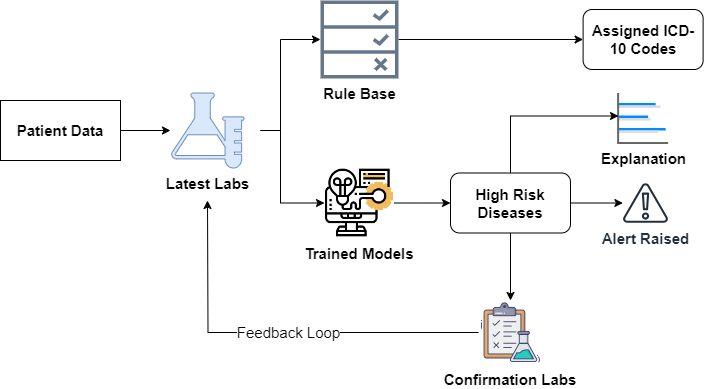}
        \caption{Proposed Clinical Decision Support System}
        \label{fig:proposed_cdss}
    \end{minipage}\hfill
    \begin{minipage}{0.48\textwidth}
        \centering
        \includegraphics[width=\linewidth]{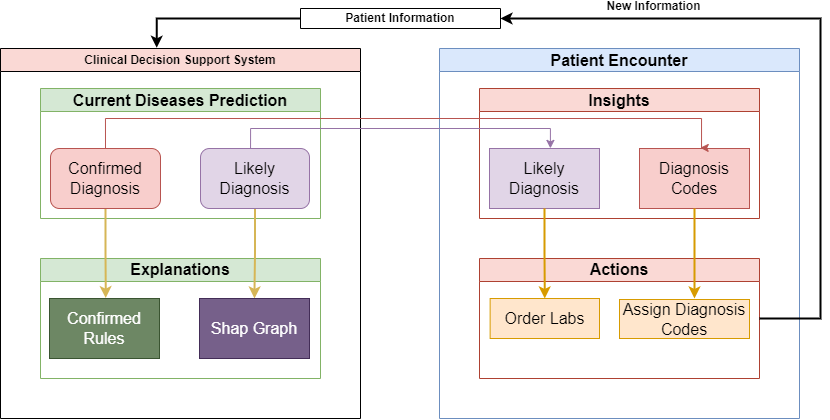} 
        \caption{CDSS Workflow}
        \label{fig:cdss_workflow}
    \end{minipage}
\end{figure}

This paper introduces a CDSS that integrates Machine Learning (ML) with a rule-based framework to provide a robust support mechanism for diagnosing a wide range of diseases with a focus on interpretability of the system shown in Figure \ref{fig:proposed_cdss}. It has the following components.

\subsubsection{Diagnosis Confirming Module}
This component assigns International Classification of Diseases (ICD-10) codes to patients based on a disease specific rule base, made using clinical guidelines of relevant associations or institutes, by facotring in demographics and and laboratory results.

\subsubsection{Likely Diagnosis Assistive Module}
By analyzing laboratory results, which offer a quantifiable snapshot of a patient's health profile, this component predicts potential diagnoses and helps physicians in focusing their diagnostic efforts.  D. Jing and M.Jing have demonstrated the potential for predicting diagnosis using laboratory results with the help of multi-class classification, achieving an F1 score of 0.76 \cite{park2021development}. Using multi-class classification on grouped diseases allows physicians to maintain focus during the process of diagnosis confirmation among a group of likely diagnoses options. For diseases that are deemed highly likely, the system also recommends further investigations to confirm these diagnoses, ensuring that the new information aids in patient centered decision-making by complying with the clinical guidelines.

\subsection*{CDSS Output}
\begin{enumerate}
    \item \textbf{Rule-Based Diagnosis Confirmation.} 
    The CDSS confirms diagnosis based on a predefined rule set that is applied to lab results and outputs ICD-10 codes.
    \item \textbf{Multiclass Probabilities (Likely Diagnosis).} The system utilizes an XGBoost multi-class classifier to compute the probability of various potential diagnoses based on lab data and presents a ranked list of likely diagnoses to clinicians.
    \item \textbf{Prediction Explanation via SHAP Values.} By quantifying the impact of each feature on the inference, SHAP values provide clinicians with explainable insights into rationale behind diagnostic inferences \cite{vimbi2024interpreting} and helps develop trust, essential for mainstream adoption of the ML based CDSS systems \cite{jones2021enhancing}
    \item \textbf{Recommendation for Follow-Up Labs.} When a likely diagnosis is identified, the system suggests follow-up investigations that could confirm the diagnosis by reducing misdiagnosis rate.
\end{enumerate}

\section{Data Overview}\label{data}
The dataset utilized for developing and validating the proposed Clinical Decision Support System (CDSS) uses demographic (age and gender) of patients, and laboratory results that are extracted from anonymized and de-identified CureMD EHR data, from 593,055 electronic medical records, making it a larger dataset from studies reported earlier in the literature ranging in sample size of 50,000 to 100,000 EMR records records\cite{gokhale2023hospital}. These data sources are curated to maintain high standards of privacy and confidentiality, adhering to HIPAA guidelines. For instance, a systematic review of hospital length of stay prediction tools found median sample sizes of around 53,211 records\cite{gokhale2023hospital}.

\begin{enumerate}
     \item \textbf{Geographical and Demographic Distribution.} The data encompasses a wide distribution across different states, medical practices and demographics (age and gender) (seen in Appendix \ref{ap:percentage_of_races} \ref{ap:state_dist}). The distribution of race is similar to the actual distribution of races \cite{jones20212020} in the United States of America. This diversity mirrors the actual real world data, ensuring the system’s applicability across various healthcare settings and trustworthiness of RWE derived.
    \item \textbf{Disease Distribution.} The dataset includes a varied distribution of diseases as seen in Table \ref{tab:disease_distribution}, allowing the system to handle multiple diagnostic scenarios and adapt to a broad spectrum of medical conditions.

\end{enumerate}

To ensure a robust and relevant participant sample, eligibility criteria was established and patients who have undergone laboratory tests within the specified one-year period prior to a
potential diagnosis were selected. Patients without age and demographic information were filtered out. The first occurrence of the diagnosis was used as a reference date for training. For normal patients, labs in the latest one-year period were used.

\subsection{Data Preparation}
Apache Spark, via PySpark, was utilized to efficiently handle and process the dataset in our on premise state-of-the-art Lakehouse. Along with Lab values, age and gender were used as features since interpretation of labs requires age and gender\cite{fenton2018glomerular}\cite{suwanrungroj2024establishing}. Lab tests including CBC, CMP, lipid panel and liver function panel were also used. 
The dataset includes laboratory data collected over a 23-year period from 2000 to 2023. For the purpose of diagnosis, we consider only the laboratory data from a one-year period immediately preceding the confirmed diagnosis of a disease. This time frame is selected to capture the most relevant biochemical changes that might indicate the onset of a disease. This ensures that the lab results added are recent and relevant \cite{wen2022clinlabomics}\cite{dimeski2022biochemical}. Normal patients are selected on the basis that they have not been diagnosed with any of the ICD-10 codes added in the pool of diseases. ICD-10 codes for classification were chosen and grouped together as shown in Table \ref{tab:disease_distribution} based on the feedback from medical professionals, leveraging expert clinical insights to inform the modeling process. Outliers were identified and removed based on standard deviation to clean the clinical errors in data reporting, which was verified with the help of visual tools such as violin plots. 

Lab names were initially mapped using IMO API to remove any errors in them, which were further refined by mapping of same lab tests onto one key. This ensured that a specific lab was not repeated in the pool with a little different name. After mapping of lab names, the lab units were mapped. It was made sure that the units were consistent. The values with inconsistent units were ignored. The mappings of both lab names and lab units were verified by clinical experts.

\begin{table}
    \centering
\caption{Disease Distribution}
\label{tab:disease_distribution}
    \begin{tabular}{|>{\centering\arraybackslash}p{0.35\linewidth}|>{\centering\arraybackslash}p{0.25\linewidth}|>{\centering\arraybackslash}p{0.15\linewidth}|} \hline 
         \textbf{Disease}&  \textbf{ICD 10 Codes} &\textbf{Count}\\ \hline 
         Upper Respiratory Tract Infections (URTI)&  J02, J03, J32, J31, J01, J06, J00&102,802\\ \hline 
         Gastroesophageal Reflux Disease (GERD)&  K29, K21, K30, K20, K25&50,086\\ \hline 
         Lung diseases&  J45, J44, J43, J42, J40, J20&49,121\\ \hline 
         Type 2 diabetes mellitus (T2DM)&  E11&30,403\\ \hline 
         Anemia&  D64, D63, D50&60,784\\ \hline 
         Kidney diseases&  N18, N17&25,553\\ \hline 
         Hypothyroidism&  E00, E01, E02, E03,
&24,084\\ \hline 
         Ischemic Heart disease (IHD)&  I20, I21, I25&15,553\\ \hline 
         Dyslipidemia&  E55&77,558\\ \hline 
 Vitamin D Deficiency& E78&69,122\\ \hline
 Disorders of White Blood Cells& D70, D71, D72, D76&15,346\\\hline
 Normal& -  -  -  -&72,643\\\hline
    \end{tabular}   
\end{table}

\section{Methodology}
\subsection{Diagnosis Confirmation Methodology}
The Diagnosis Confirmation System is designed to utilize lab results to confirm patient diagnoses. It is based on a rule-based expert system, where each rule within the rule base can confirm a diagnosis, if conditions in the antecedents are met. These rules and conditions were extracted by a panel of physicians from the clinical guidelines. The rules look for anomalies in the lab reports according to the clinical guidelines, and then confirm a diagnosis.

\subsubsection{Rule base Design:}
Each rule in the rule base is comprised of multiple conditions that must all be true for the rule to confirm a diagnosis. Each condition is comprised of:
\begin{itemize}
    \item The lab test identifier
    \item The comparison value
    \item The unit of measurement
    \item The type of comparison (e.g., greater than, less than, equal to)
\end{itemize}
Currently, the system includes rules for 59 health conditions, and we want to expand by incorporating rules from guidelines, extracted by our team of clinical analysts. The patient lab results are evaluated against the stored rules to confirm a diagnosis. This process is fully automated and provides a robust framework for assigning ICD-10 codes based on the latest lab data.

\subsection{Likely Diagnosis Methodology}
Recognizing that all labs results might not be available, we train AI/ML models to infer the set of likely diagnoses based on available information without using any imputation. XGBoost was used for multi-class classification since it inherently manages missing values \cite{chen2016xgboost}. It is also explainable using SHAP \cite{li2022extracting}, and outperforms the bagging ensembles\cite{ogunleye2019xgboost}.



\subsubsection{Data Utilization and Partitioning:}

80\% of the total dataset was allocated for training and internal validation purposes. Stratified splitting approach was used to ensure that each split is representative of the overall population distribution. Cross validation was performed using 5-fold cross-validation. The remaining 20\% served as a standalone test set to evaluate the model’s performance. Grid search cross-validation approach was used to tune hyper-parameters including \texttt{max\_depth}, \texttt{learning\_rate}, \texttt{n\_estimators}, \texttt{gamma}, \texttt{subsample}, \texttt{colsample\_bytree}, and \texttt{reg\_lambda}.



\subsubsection{Class Imbalance}

During experimentation, classes were balanced by under sampling to the count of the smallest disease group. This resulted in a loss of data and decrease in performance of our model. Then weights for each class were initialized based on the inverse frequency of the disease in the training data. Through empirical studies, a heuristic was found: taking the square root of these weights yielded the best results. This approach helped to enhance model sensitivity for less prevalent conditions without compromising the overall accuracy of the inference system.




\section{Results and Discussion}

We used the Top N criterion (Most probable N diseases) from \cite{park2021development} to evaluate the accuracy of our likely diagnosis multi-class model. In this criterion, we check if the correctly diagnosed disease is within the N highest probability diseases. We also used confusion matrices to check the precision of our models. We then compared the distribution of our predictions among diseases to the true distribution of diseases in the EMR data.
\begin{table}[h]
\centering
\begin{tabular}{|>{\centering\arraybackslash}m{0.1\linewidth}|c|c|c|c|c|c|c|c|c|c|c|c|c|c|c|c|}
\hline
\textbf{Top-N} & 1 & 2 & 3 & 4 & 5 & 6 & 7 & 8 & 9 & 10 & 11 \\ \hline
\textbf{Accuracy}& 31.18 & 52.65 & 66.43 & 76.00 & 83.10 & 88.43 & 92.43 & 95.40 & 97.49 & 98.87 & 99.6 \\ \hline
\end{tabular}
\caption{Top-N Accuracy (\%)}
\label{tab:top_n_accuracy_recall}
\end{table}

\vspace{-40pt} 
\subsection{Accuracy and Recall}
The table \ref{tab:top_n_accuracy_recall} shows the accuracy results of the model for different values of N. As it can be seen from the table \ref{tab:top_n_accuracy_recall}, the 80 percent threshold is achieved when the Top 5 approach is used. This 80 percent threshold serves as an optimal trade off between the number of predicted diseases and an increase in accuracy if the disease set is increased after 5. To see the recall of individual diseases, Table \ref{tab:recall_rates} shows the results. Disease groups like URTI, Dyslipidemia and Anemia are predicted in top 5 with a recall of 0.943, 0.866, 0.860 respectively. Even in Top 1 results, The best performance of the model is in distinguishing the disease patients from the normal patients with a very high recall of 0.91 in Top 5 diseases. The Top 5 approach is being considered here since our goal is to predict the likely diagnoses, recommend the subsequent lab investigations, and confirm it using our rule base.

\begin{table}[ht]
\centering
\caption{Recall rates for various diseases at different rankings}
\label{tab:recall_rates}
\begin{tabular}{|l|r|r|r|r|r|} \hline 

\textbf{Disease} & \textbf{Top 1} & \textbf{Top 2} & \textbf{Top 3} & \textbf{Top 4} & \textbf{Top 5} \\ \hline 

URTI & 0.332 & 0.629 & 0.810 & 0.902 & 0.943 \\ \hline  
GERD & 0.235 & 0.384 & 0.504 & 0.659 & 0.802 \\ \hline  
Lung\_Disease & 0.124 & 0.289 & 0.500 & 0.686 & 0.827 \\ \hline  
T2DM & 0.202 & 0.356 & 0.493 & 0.596 & 0.685 \\ \hline  
Anemia & 0.429 & 0.639 & 0.758 & 0.825 & 0.860 \\ \hline  
KD & 0.359 & 0.599 & 0.708 & 0.776 & 0.815 \\ \hline  
Hypothyroidism & 0.171 & 0.279 & 0.375 & 0.451 & 0.528 \\ \hline  
IHD & 0.126 & 0.213 & 0.308 & 0.397 & 0.488 \\ \hline  
Dyslipidemia & 0.303 & 0.541 & 0.688 & 0.788 & 0.866 \\ \hline  
V\_Ddeficiency & 0.310 & 0.542 & 0.668 & 0.752 & 0.823 \\ \hline  
D\_WBC & 0.297 & 0.454 & 0.552 & 0.620 & 0.674 \\ \hline 
normal & 0.495 & 0.726 & 0.824 & 0.875 & 0.916 \\ \hline

\end{tabular}
\end{table}

\begin{figure}
    \centering
    \includegraphics[width=1\linewidth]{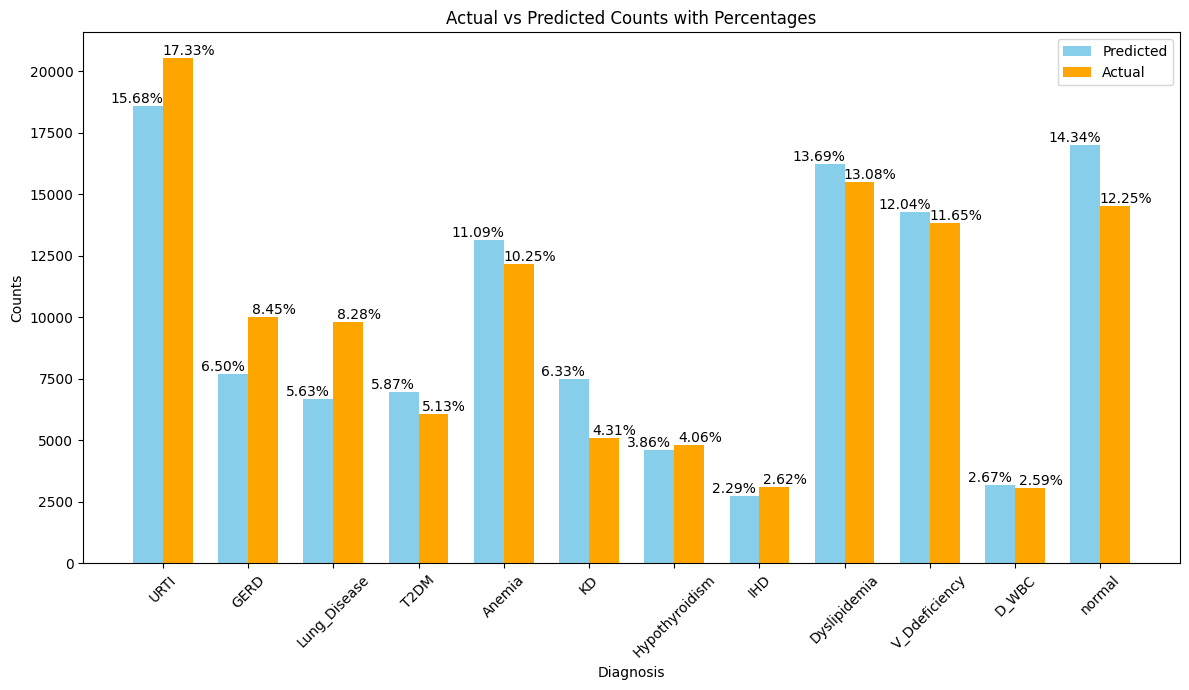}
    \caption{Actual vs Predicted Counts with Percentages}
    \label{fig:pred_percentage}
\end{figure}

\subsection{Confusion Matrices}
\begin{figure}
    \centering
    \includegraphics[width=0.8\linewidth]{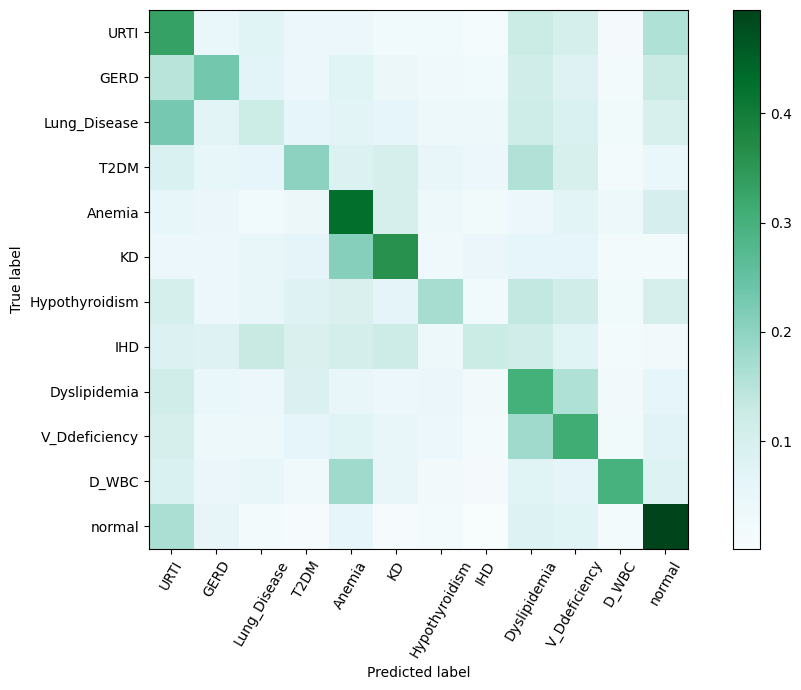}
    \caption{Precision in disease forecasting}
    \label{fig:cm1}
\end{figure}

The precision for the multi-class model trained for likely diagnosis is presented in figure \ref{fig:cm1}. It shows the density of a disease being predicted correctly. This helps to observe what label is being predicted for which disease. It can be observed that the label 'normal' has the best results in the sense of being predicted as normal. Furthermore, the color density shows that predicted label is the same as the true label for most of diseases in the disease set, and is evident from the color density in the diagonal. It can be seen that URTI patients are being classified as lower lung disease patients, which makes sense as both disease groups are related to lungs and have similar set of symptoms. For other labels, the model is very precise which can be seen from the color intensity in the center of the graph.


\subsection{Prediction Distribution}
Figure \ref{fig:pred_percentage} shows the distribution of predicted labels compared to the true labels. When considering the Top 1, the predictions of our model matches the true prevalence of these diseases in the EMR data very closely. For example, 15.68\% patients (18.5k approx) in the test data have URTI, and our model matches it very closely and predicts prevalence of URTI for 17.33\% patients. This allows us to ensure that our model is inferring the correct group of diseases amongst the different disease groups, and is not over-predicting or under-predicting for any particular disease group. The distribution of weights while training the model helped bring the true vs predicted counts closer to one another. The maximum difference can be seen in the case of Lung\_Disease which is lower than 3\%.

\section{Conclusion}
In this paper, we propose a novel disease inference engine that fuses the knowledge from a rule based expert system with the prediction models of ML to overcome the shortcomings of each one of them. The likely diagnosis system serves as an assistant to a physician, helping him in correctly diagnosing a patient and subsequent management according to the guidelines of respective institutes and associations. The physician can then order further lab investigations to confirm diagnoses in the likely set of diagnoses. When the lab results are received in the EMR system, the expert system confirms the diagnosis based on the rules.

For making the sytem explainable to a physician, the mean Shap (Appendix \ref{ap:shap_plots}) were used and learned trends were validated. The Mean Shap values show the overall Trend learned by the model while the Shap values for individual patient data (Appendix \ref{ap:risk_features}) help understand the model decision for a patient. This provides a physician with a reasoning of why the inference model is giving a certain output and can gain the trust of a physician, if the explanations are grounded in medical knowledge.

By optimizing the diagnosis accuracy and reducing the redundancy of unnecessary tests, the proposed system can become a componnent of CDSS, and can significantly enhance its operational efficiency by reducing healthcare costs. It's ability to handle multiple diseases and integrate new information makes it a robust system that can adapt to various clinical needs, thus broadening its applicability and generalizability.

\subsection*{Limitations and Future Work}

The study, while robust in its achievements, faces several limitations that could affect its application and generalizability. The scope of the research is limited to using laboratory data as the primary source of information for making predictions. While lab results are helpful for many conditions, the exclusion of other critical factors such as vital signs, social history, and patient-reported symptoms could limit the comprehensiveness and accuracy of the predictions. The reliance solely on the lab investigations might introduce biases where certain diseases that manifest predominantly through other clinical signs or patient histories are underrepresented or inaccurately predicted.

The future direction of research includes using lab specific data in a time window in which lab results are more meaningful. Beyond recommending the follow-up labs, the system will be enhanced to include include recommendations for medications, procedures, surgeries, and treatment pathways. This would assist in managing patients using a systems medicine approach.

\bibliographystyle{plain}
\bibliography{references}


\newpage
\pagenumbering{Alph} 

\appendix
\section{Data Distributions}\label{ap:data_distribution}
\subsection{Race Distribution}

\begin{figure}[ht!]
    \centering
    \includegraphics[width=0.8\linewidth]{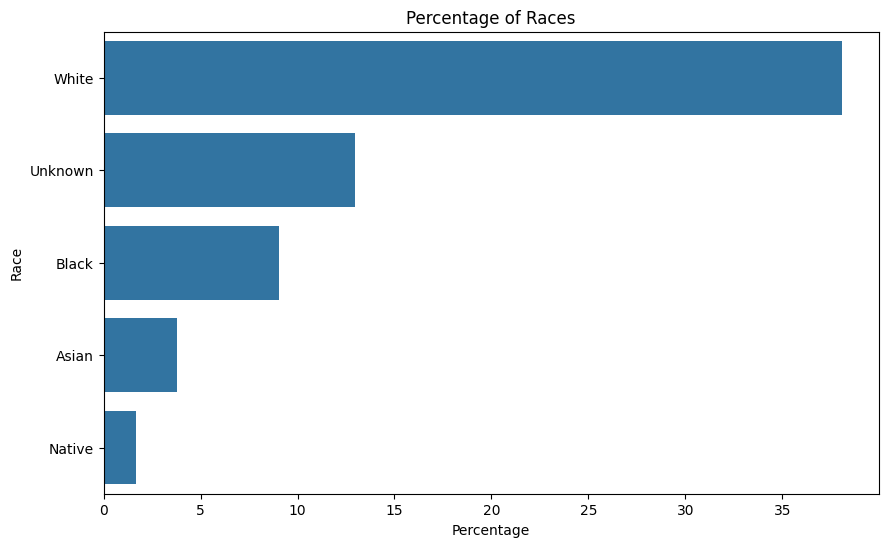}
    \caption{Race Distribution of Data}
    \label{ap:percentage_of_races}
\end{figure}

\subsection{Statewise Data Distribution}

\begin{figure}[ht!]
    \centering
    \includegraphics[width=0.8\linewidth]{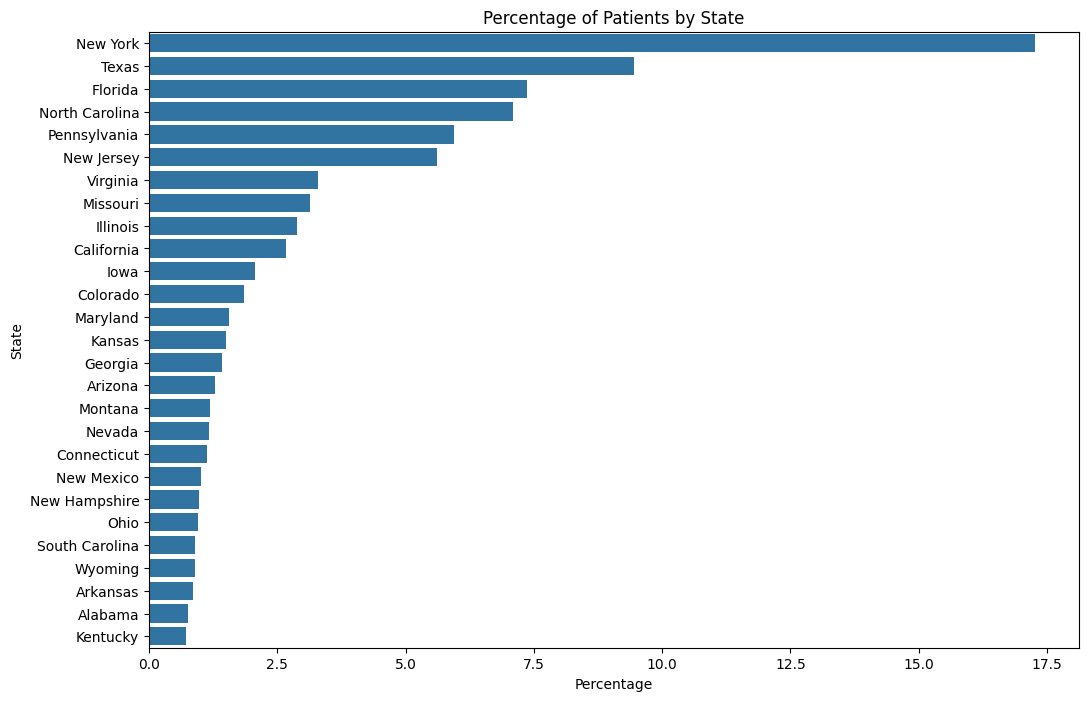}
    \caption{Distribution of Patients by States}
    \label{ap:state_dist}
\end{figure}

\section{Shap Plots} \label{ap:shap_plots}

\begin{figure}[htbp]
    \centering
    \includegraphics[width=1\linewidth]{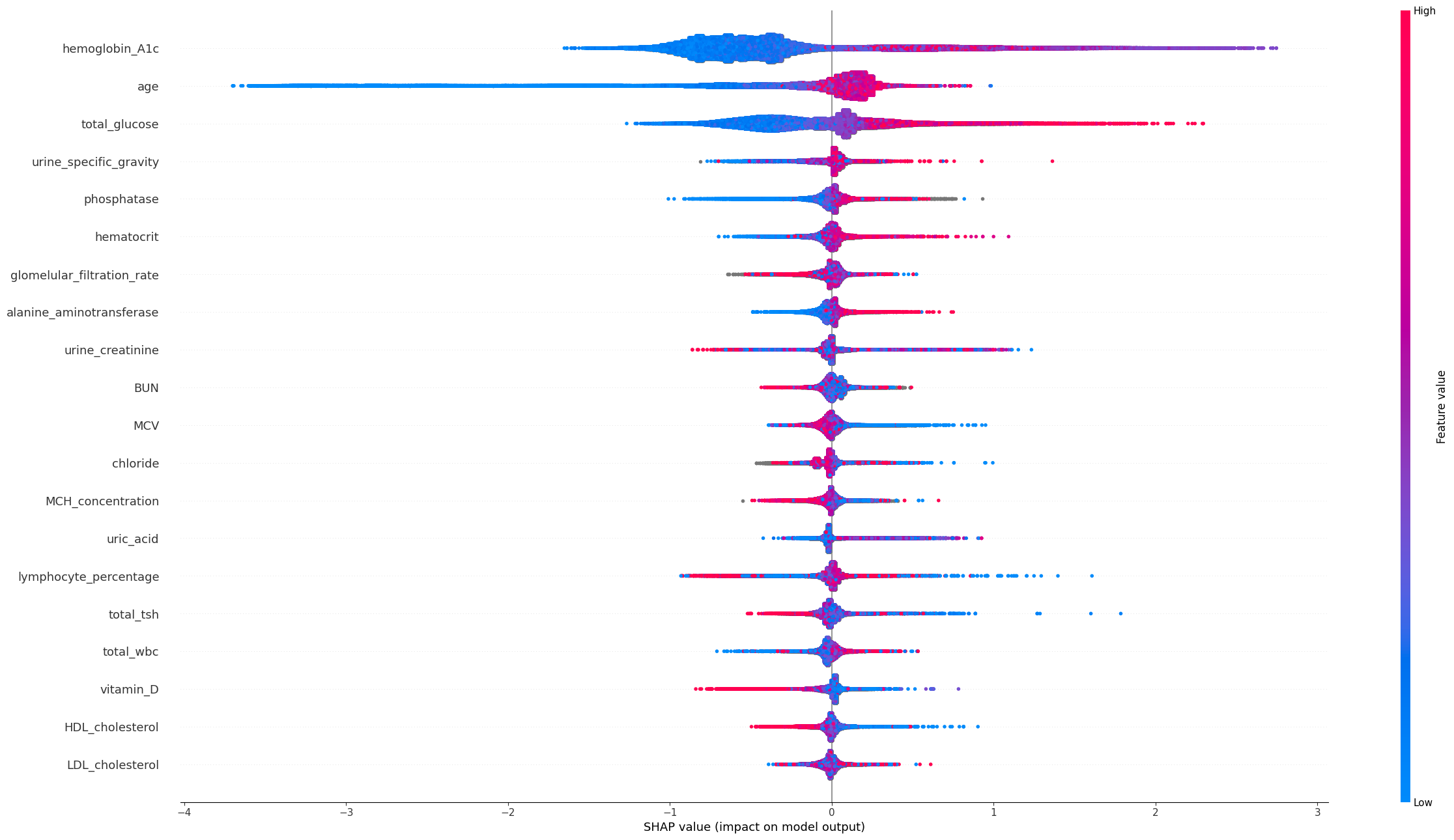}
    \caption{Shap summary plot for T2DM}
    \label{ap:shap_T2DM}
\end{figure}
\begin{figure}[htbp]
    \centering
    \includegraphics[width=0.5\linewidth]{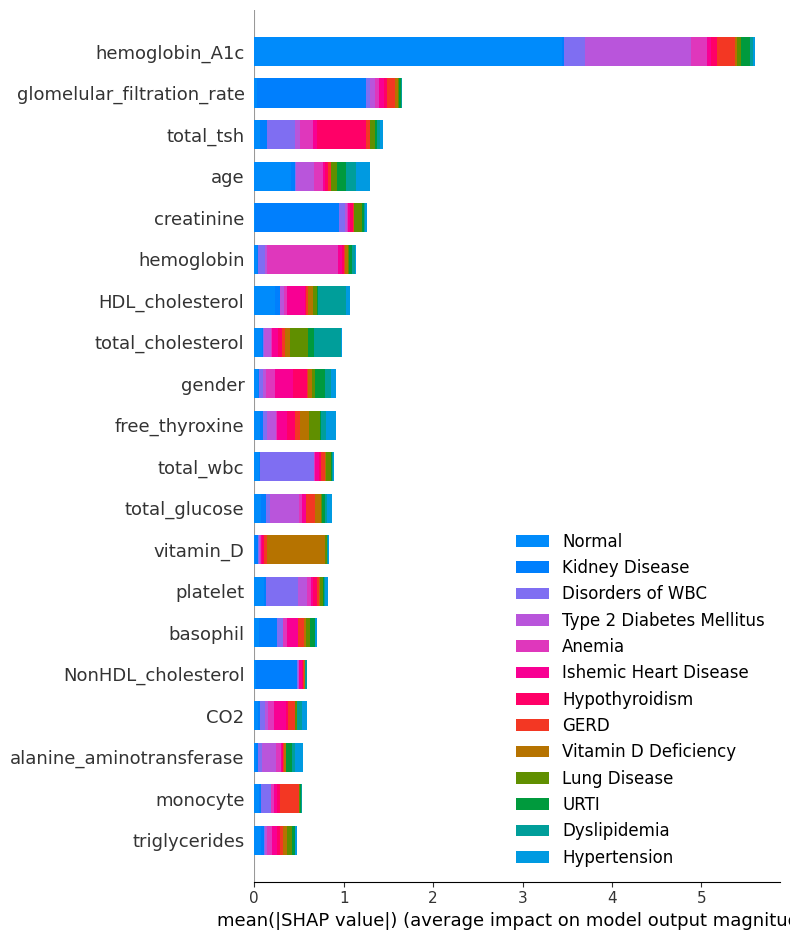}
    \caption{Summary plot showing absolute impact of feature on decision}
    \label{ap:shap_summary_patient}
\end{figure}
\begin{figure}[htbp]
    \centering
    \includegraphics[width=1\linewidth]{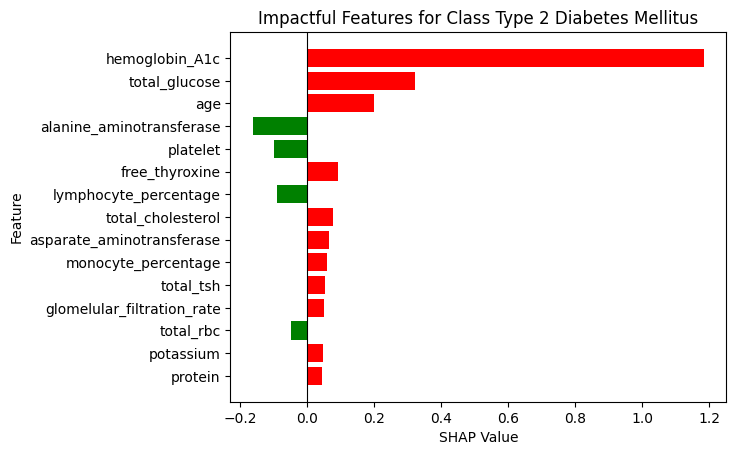}
    \caption{Important risk features and their impact for a given patient}
    \label{ap:risk_features}
\end{figure}

\end{document}